\documentclass{article}

\usepackage[T1]{fontenc}
\usepackage[preprint]{corl_2026} 
\usepackage{amsfonts}       
\usepackage{amsmath}
\usepackage{natbib}
\usepackage{graphicx}
\usepackage{wrapfig}
\usepackage{cleveref}
\usepackage{algorithm}
\usepackage[endLComment=,italicComments=false]{algpseudocodex}
\usepackage{enumitem}
\usepackage{caption}

\usepackage{colortbl} 
\definecolor{myred}{HTML}{F54254}
\definecolor{myorange}{HTML}{FFB135}
\definecolor{mygreen}{HTML}{10BD35}
\definecolor{myblue}{HTML}{598BE7}
\definecolor{mypurple}{HTML}{9A1C6B}
\definecolor{plgray}{HTML}{999999}

\renewcommand{\phi}{\varphi}

\title{FlashRFCL: Demonstration-Efficient Sparse Reward RL via Reverse-Forward Curriculum Learning in High-Throughput GPU Simulation}

\author{
  Yong Woo Choi \\
  UC San Diego \\
  \And
  Stone Tao \\
  UC San Diego \\
  \And
  Henrik I. Christensen \\
  UC San Diego \\
}

\begin{document}
\maketitle


\begin{abstract}
    Solving complex tasks using reinforcement learning (RL) from sparse rewards and demonstrations is appealing since one can avoid reward shaping. However, past approaches often require heavy curriculum engineering, extensive training time, and/or infeasible amounts of demonstrations. In this paper, we investigate how scalable RL algorithms, high-throughput GPU simulations, and demonstration-based auto-curricula can address these challenges. We show that our method, FlashRFCL, can solve a variety of high-dimensional robotic tasks faster in addition to relying on just a single demonstration each. We perform extensive ablation studies on the design choices of FlashRFCL, highlighting how each component impacts training speed and performance compared to prior work.
\end{abstract}



\section{Introduction}
Reinforcement learning (RL) has shown success in robot learning for training models to high success and low cycle times, whether it is via fine-tuning Vision-Language-Action models \citep{DBLP:journals/corr/abs-2511-14759} or RL followed by sim-to-real transfer. While imitation learning remains popular, scaling RL sim-to-real does not require massive demonstration datasets. Instead, with appropriate modelling of a task and dense reward shaping scaling RL has led to successes in real-world locomotion~\citep{seo2025learningsimtorealhumanoidlocomotion, rudin2022learningwalkminutesusing, tairanhe2025} and manipulation~\citep{yin2026omnireset, dextrahrgb, almuzairee2026squint, xie2026handelbotrealworldpianoplaying, fang2026dexdrummer}. Many of the recent sim-to-real advances can be attributed to the high-throughput data collection speeds of GPU-parallelized simulators such as Isaac~\citep{makoviychuk2021isaac} and Mujoco Warp~\citep{todorov2012mujoco}.

Although GPU-parallelized simulation enables inexpensive large-scale data generation, many existing approaches still rely on carefully shaped dense rewards for each task and can easily lead to suboptimal policies due to reward hacking~\citep{DBLP:journals/corr/AmodeiOSCSM16}. Training with sparse rewards provides a more scalable alternative to dense reward shaping, but often requires prohibitively large amounts of exploration and long training times for tasks of practical interest. To overcome the exploration problem in sparse reward settings, online learning from demonstrations (LfD) is a promising direction to guide RL exploration toward successful behaviors. Since collecting large demonstration datasets is infeasible for complex robotics tasks, methods that can generalize from limited demonstrations to cover the task's behavior space are highly desirable. Inverse reward learning methods such as GAIL~\citep{gail2016} and different flavors of off-policy RL with demonstrations~\citep{lancaster2023modem, ball2023efficientonlinereinforcementlearning, mu2024drs} fall under this category. However these methods tend to be unstable or require very long training times.

In this work, we introduce FlashRFCL, a fast and demonstration efficient LfD algorithm that leverages aggressive state-reset based curricula and large-scale GPU simulation. Specifically, we extend and adapt Reverse Forward Curriculum Learning (RFCL)~\citep{tao2024rfcl} for GPU simulation to create our FlashRFCL algorithm. We swap out the base RL algorithm in RFCL from SAC to FlashSAC~\citep{kim2026flashsacfaststableoffpolicy}, and introduce key changes to hyperparameters and curriculum design to better suit the sparse reward and demonstrations setting. Through experiments in simulated robot manipulation tasks, we demonstrate that FlashRFCL consistently outperforms baselines in wall time and demonstration efficiency. We further perform analysis into how FlashRFCL scales to better understand how exactly the massive parallel data collection regime impacts training. Our ablations further demonstrate how our curriculum design changes improve the robustness of FlashRFCL to the number of parallel environments, making it performant in both low and high-parallelism settings.


\section{Related Work}
\label{sec:related-work}
\textbf{Wall-Time Efficient Reinforcement Learning Algorithms:}
In massively parallel simulation settings where environment interactions are relatively inexpensive, the primary focus naturally shifts from sample efficiency to wall-clock training time. In this regime, on-policy RL algorithms such as PPO have achieved widespread success across domains including games and robotics, largely due to their stability and ease of scaling across distributed environments~\citep{openai2019dota2largescale, openai2019solvingrubikscuberobot, rudin2022learningwalkminutesusing}. However, on-policy RL algorithms are inherently sample-inefficient as they require data to be collected from the current policy and discard past experiences after each update. In contrast, off-policy RL algorithms can leverage heterogeneous data sources, such as expert demonstrations or trajectories from prior policies, enabling significantly greater data reuse. Despite this advantage, off-policy RL algorithms are rarely used in massively parallel control tasks. Recent work, however, has shown that with improved algorithmic design, off-policy RL algorithms can scale to match or even exceed the final performance of PPO while achieving faster convergence in wall-clock time~\citep{kim2026flashsacfaststableoffpolicy, seo2025learningsimtorealhumanoidlocomotion, seo2025fasttd3, li2023parallelqlearningscalingoffpolicy, raffin2025isaacsim, shukla2025fastsac}. Similar to past work we use off-policy RL in GPU simulations, with the key difference being that we tackle sparse reward problems and leverage demonstration data.

\textbf{Curriculum Learning:}
Curriculum learning is a training strategy that presents training examples in a structured order of increasing difficulty, enabling faster and more stable learning \citep{10.1145/1553374.1553380, jiang2021prioritizedlevelreplay, parkerholder2023evolvingcurricularegretbasedenvironment}. Simulation enables curriculum learning through direct resets to states that are difficult to reach under the current policy but useful for learning. Prior work has leveraged this through state-reset curricula, where the initial state distribution is adapted during training to improve exploration. Some approaches uniformly reset to states from demonstrations ~\citep{nair2018overcomingexplorationreinforcementlearning} or past experiences ~\citep{goexplore}. Reverse curricula are another approach where policies start learning from states near goals and work backwards \citep{florensa2018reversecurriculumgenerationreinforcement}. Our work is an extension of Reverse Forward Curriculum Learning (RFCL)~\citep{tao2024rfcl} that tackles sparse reward robotics tasks with few demonstrations and constructs automatic curricula out of demonstration states. We extend and adapt the RFCL algorithm to the massively parallel regime and analyze how design choices in curricula and RL impact performance. Our contributions significantly improve both the wall-time and demonstration efficiency compared to RFCL.
    

\section{Preliminaries}
\label{sec:preliminaries}

\subsection{Reinforcement Learning}

We study reinforcement learning in a Markov Decision Process (MDP) defined by the tuple $(\mathcal{S}, \mathcal{A}, p, r, \gamma)$ where $\mathcal{S}$ is the state space, $\mathcal{A}$ is the action space, $p(s' \mid s, a)$ denotes the (unknown) transition dynamics, $r(s, a)$ is the reward function, and $\gamma \in [0, 1]$ is the discount factor. The objective is to learn a policy $\pi_{\theta}(a|s)$ parameterized by $\theta$ that maximizes the expected discounted return, $\max_\theta \mathbb{E}_{\pi_\theta}[\sum_{t=0}^{\infty} \gamma^{t}r_{t}]$. In this work, we focus on sparse reward functions, where $r(s, a) = 1$ if the next state is a success state and $0$ otherwise. To mitigate exploration challenges arising from sparse rewards in high-dimensional tasks, we assume access to a dataset of $N$ demonstrations $\mathcal{D}_{demo} = \{\tau_1, ..., \tau_N\}$, where each demonstration $\tau_i = \{s_{i,0}, a_{i,0}, ..., s_{i, T_i-1}, a_{i, T_i-1}, s_{i,T_i}\}$ is a trajectory of length $T_i$ consisting of a sequence of states and actions.

\subsection{Reverse Forward Curriculum Learning (RFCL)}

\begin{wrapfigure}{r}{0.5\linewidth}
    \centering
    \small
    \vspace{-2.5em}
    \includegraphics[width=\linewidth]{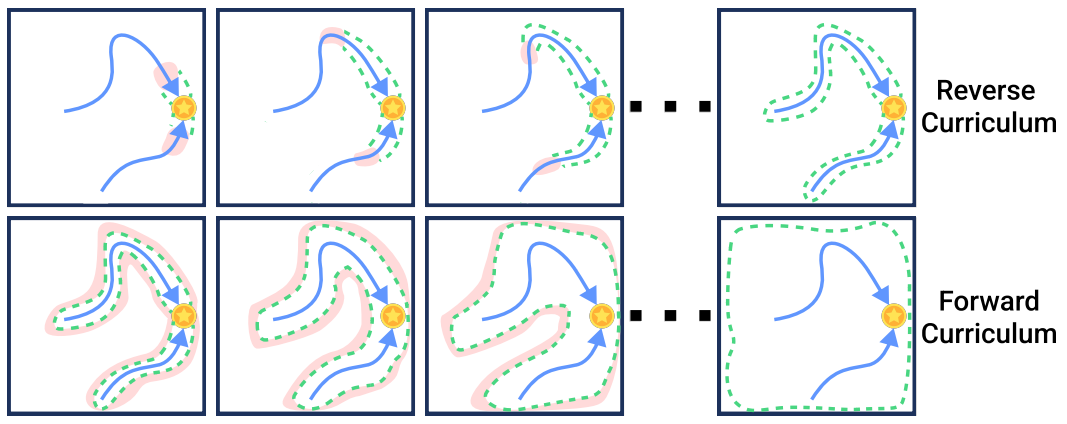}
    \vspace{-10pt}
    \caption{Original figure from RFCL. Visualization of reverse and forward curriculum progression. Blue arrows denote demonstration trajectories leading to the goal (gold star). The dashed green region represents states from which the policy achieves high return, and the red shaded region represents the most frequently sampled initial states at each stage of the curriculum.}
    \label{fig:rfcl}
    \vspace{-2.5em}
\end{wrapfigure}

Our work is an extension of RFCL adapted to support both the low and high-parallelism data regimes of RL. RFCL is a two-stage training scheme consisting of a reverse curriculum followed by a forward curriculum. The reverse curriculum first learns a policy that performs well near states observed in the demonstrations. The forward curriculum then generalizes this policy to the true initial state distribution $\mathcal{S}_{init}$ by prioritizing initial states with higher estimated learning potential. This section re-describes RFCL and defines key terms used throughout the paper, with a visualization of the curriculum design shown in Fig. \ref{fig:rfcl}.

\textbf{Reverse Curriculum:}  At the start of training, a reverse curriculum is constructed for each demonstration $\tau_i$ by resetting the environment to states along the demonstration. Specifically, the reverse curriculum for $\tau_i$ is defined as a sequence of stages $(\rho_{T_i}^{r}, \rho_{T_i-\delta}^{r}, ..., \rho_{0}^{r})$ for a small reverse step size $\delta$, $r$ here refers to "reverse". Each stage $\rho_{t}^{r}$ defines a sampling distribution over demonstration states $s_{i, t+k}$, with $k \sim K$ and $k \in \{0, 1, ..., K_{\max}\}$ for some maximum discrete offset $K_{\max}$. RFCL adopts a geometric distribution for $K$. 

At the beginning of each episode, RFCL samples a demonstration $\tau_i$ and an offset $k$, and reset the environment to state $s_{i, t+k}$. The reverse curriculum for $\tau_i$ advances to the next stage when the policy achieves success in the last $m$ episodes initialized at state $s_{i, t}$. When $\rho_0^r$ is reached and the policy achieves a high success rate, RFCL labels demonstration $\tau_i$ as being reverse-solved.

\textbf{Forward Curriculum:} After all demonstrations are reverse-solved, RFCL uniformly sample a set of $n$ initial states from $\mathcal{S}_{\text{init}}$. These initial states are combined with the demonstration initial states $\Lambda_{\text{demo}}$ to form a train set of initial states $\Lambda_{\text{train}} = \{s_{i, \text{init}}\}_{i=1}^n$ for the forward curriculum. The forward curriculum is defined as a sequence of stages $(\rho_{0}^{f}, \rho_{1}^{f}, ...)$, where each $\rho_{i}^{f}$ defines a sampling distribution over $\Lambda_{\text{train}}$ as a convex combination of a score-prioritized distribution $P_S$ and a staleness-prioritized distribution $P_C$: $P_{\rho_i^f} = (1 - \rho) \cdot P_S + \rho \cdot P_C$
where $\rho \in [0, 1]$ is a staleness coefficient. 

Following PLR \citep{jiang2021prioritizedlevelreplay}, RFCL adopts a rank-based prioritization scheme for $P_S$. Let $q$ denote the fraction of successful episodes among the last $k$ episodes initialized from $s_{i, \text{init}}$. For a success-rate threshold $\omega \in [0, 1]$, RFCL uses a simple score function that assigns a score of $2$ if $q = 0$, a score of $3$ if $0 < q < \omega$, and a score of $1$ if $q \geq \omega$. Given scores $S = \{S_i\}_{i=1}^n$ over $\Lambda_{\text{train}}$, initial states are ranked according to $S$, which are then used to define the score-prioritized distribution:
\begin{equation}
    P_S(s_{i, \text{init}}|\Lambda_{\text{train}}, S) = \frac{\text{rank}(S_i)^{-1/\beta}}{\sum_{j} \text{rank}(S_j)^{-1/\beta}}
\end{equation}
where $\beta$ is a temperature parameter controlling the degree of prioritization. The staleness-prioritized distribution $P_C$ is defined analogously based on staleness scores $C = \{C_i\}_{i=1}^n$:
\begin{equation}
    P_C(s_{i, \text{init}}|\Lambda_{\text{train}}, C, c) = \frac{c - C_i}{\sum_{C_j \in C} c - C_j}
\end{equation}
where $C_i$ is the episode count at which initial state $s_{i, \text{init}}$ was last sampled and $c$ is the total number of episodes rolled out in training.
	
\section{Method}
\label{sec:method}
\subsection{Scaling and Robustifying RFCL}
    \textbf{Reinforcement Learning:} FlashRFCL extends RFCL by adapting to massively parallelized simulation regimes to accelerate learning in complex environments from sparse rewards. As the base RL algorithm, we swap out RFCL's use of SAC~\citep{haarnoja2018softactorcriticoffpolicymaximum} with FlashSAC~\citep{kim2026flashsacfaststableoffpolicy} which is better suited for GPU parallelized simulations. Notably, we adopt key changes to SAC from FlashSAC such as large-batch updates, distributional critics, and n-step returns. Different to RFCL, the separate offline demonstration replay buffer containing demonstration transitions is optional in our method. Our experiments in Section \ref{sec:train_without_demo_actions} demonstrate how FlashRFCL can still solve tasks using the demonstration simulation states alone without needing action labels and a offline buffer.

    \textbf{Update-To-Data Ratio:} 
    FlashSAC achieves faster convergence by scaling model and batch sizes while reducing the update-to-data (UTD) ratio. However, in our experiments we find that low UTD ratios result in reduced wall-clock efficiency, and therefore adopt a higher UTD ratio. One reason for this discrepancy may arise from differences between sparse- and dense-reward training dynamics where more gradient updates are needed for accurate critics in low-signal settings like sparse reward tasks. Another reason is that our evaluation focuses mostly on manipulation tasks, which prior work suggests may benefit more from higher UTD ratios~\citep{seo2025fasttd3}.

    \textbf{Curriculum Modifications:}
    One drawback of RFCL is that reverse curriculum learning can stall training when a particular demonstration fails to be reverse-solved, preventing transition to the forward curriculum. The motivation for a two stage split in RFCL between reverse and forward curricula was to ensure after the reverse curriculum stage RFCL would have performed enough exploration to train a somewhat robust policy. However, in GPU simulations, we find that exploration challenges caused by limited demonstrations are alleviated. Motivated by this, we construct a smoother curriculum that starts the forward curriculum once a single demonstration is reverse-solved, enabling faster expansion of state-action coverage from already reverse-solved demonstrations.
    
    Let $\alpha$ denote the online/offline sampling ratio and let $p_{\text{reverse}}$ denote the fraction of demonstrations that are not reverse-solved. For each new episode, we draw a Bernoulli variable with probability $p_{\text{reverse}}$ to decide whether to initialize from a demonstration state (reverse curriculum) or from an initial state sampled from $\Lambda_{\text{train}}$ (forward curriculum). Similarly, the probability of sampling transitions from the offline replay buffer is annealed according to $\alpha \cdot p_{\text{reverse}}$. We fix $\alpha = 0.5$ across all tasks. Consequently, offline replay transitions initially constitute 50\% of training samples and are gradually phased out as the reverse curriculum completes for all demonstrations.

\subsection{Improving Demonstration Efficiency}
    When few demonstrations are provided, the policy is overfit to a narrow state-action distribution. As a result, the policy frequently yields zero return when the simulation is initialized to most initial states in $\Lambda_{\text{train}}$ during the forward curriculum. This issue is further exacerbated when using fewer environments in low parallelization regimes of RFCL, as substantially more training iterations are required to encounter states that the policy could solve the task from. To address this, we propose a simple initial-state sampling strategy that improves demonstration efficiency. We further introduce a variant of FlashRFCL with a minor algorithmic modification that can be used in conjunction with this sampling scheme, enabling effective learning from a single demonstration even with a small number of environments (e.g. 16). An algorithmic description is provided in Appendix \ref{appendix:algo}.

    \textbf{Demo-Stratified Sampling (DSS):}
        We define states that the policy $\pi_\theta$ can achieve nonzero return as boundary-states. As $\pi_\theta$ becomes proficient over a broader state-action distribution, it increasingly encounters boundary-states, reducing the difficulty of the forward curriculum. This motivates a more fine-grained sampling strategy near $\{s_{i, \text{init}}\}_{i=1}^N$ of $\Lambda_\text{demo}$ to preserve strong learning signals.

        DSS constructs $\Lambda_\text{train}$ by biasing the sampling towards $\Lambda_\text{demo}$. First, we uniformly sample $m \gg n$ candidate initial states $\Lambda_\text{cand}$ from the true initial-state distribution $\mathcal{S}_\text{init}$. For each demonstration initial state in $\Lambda_\text{demo}$, we compute the Euclidean distance to every candidate state. To ensure balanced stratification across demonstrations, distances are normalized independently per demonstration using empirical means and standard deviations to obtain z-scores. We then define a sequence of $Q$ increasing quantile levels $\{p_k\}_{k=1}^{Q}$ in the range $[0, 1]$ and compute the corresponding per-demonstration z-score quantiles $\{q_k^{(i)}\}_{i=1, k=1}^{N, Q}$. These quantiles induce strata $\{\Lambda_\text{near-demo}^{(k)}\}_{k=1}^{Q}$ over candidate states according to their normalized proximity to the demonstrations. To bias sampling toward states closer to demonstrations, we allocate target sample counts $\{N_k\}_{k=1}^{Q}$ using an exponentially decaying distribution over strata. We set $N + \sum_{k=1}^Q N_k = \left\lfloor \frac{n}{2} \right\rfloor$ for all tasks. Then, we uniformly sample $N_k$ states from each stratum and combine them to obtain $\Lambda_{\text{near-demo}}$. Finally, to ensure sufficient coverage of $S_\text{init}$, we combine $\Lambda_\text{demo}$, $\Lambda_{\text{near-demo}}$, and an additinal uniformly sampled subset of size $n - (N + \sum_{k=1}^{Q} N_k)$ from $\mathcal{S}_\text{init}$ to obtain $\Lambda_\text{train}$.
        
    \textbf{Forward Curriculum Design Choices:}
        In the massively parallel GPU simulation regime, we find that maintaining a separate offline buffer and aggressively sampling from it (as in prior RFCL variants) is less critical for stabilizing forward curriculum learning. However, in settings with limited parallelism (e.g., vision-based RL or inherently low-throughput environments), we find DSS alone is sufficient but may be sample-inefficient in some tasks. 
        
        To address this, we introduce a lightweight state-reset curriculum. The key idea is to partition $\Lambda_\text{train}$ into a feasible set $\Lambda_\text{feasible}$ and its complement $\Lambda_\text{train} \setminus \Lambda_\text{feasible}$, and to bias resets toward feasible states until the policy acquires sufficient state-action coverage to support forward curriculum expansion. We initialize the feasible set as $\Lambda_{\text{feasible}} = \Lambda_{\text{demo}} \cup \Lambda_{\text{near-demo}}^{(1)}$ which ensures that early rollouts are likely to achieve non-zero reward and provide meaningful learning signal. We sample initial states from the mixture distribution
            $P_\text{demo-expand} = p_\text{demo} \cdot P_\text{demo} + p_\text{feasible} \cdot P_\text{feasible} + p_\text{explore} \cdot P_\text{explore}$
        where $P_{\text{feasible}}$ is a convex combination of $P_S$ and $P_C$ over $\Lambda_{\text{feasible}}$ (as defined previously), and $P_{\text{demo}}$ and $P_{\text{explore}}$ denote uniform distributions over demonstration initial states and $\Lambda_{\text{train}} \setminus \Lambda_{\text{feasible}}$, respectively.
        
        Let $x = \frac{|\Lambda_{\text{feasible}}|}{ |\Lambda_{\text{train}}|}$. We adopt the following replay schedule:
        $$
        p_{\text{explore}} = 2x(1 - x), \quad
        p_{\text{demo}} = \max(0, 0.5 - x)(1 - p_{\text{explore}}), \quad
        p_{\text{feasible}} = 1 - p_{\text{demo}} - p_{\text{explore}}
        $$
        This schedule initially prioritizes demonstration and feasible states, providing stable learning signals in the early phase. As the feasible set expands, sampling gradually shifts away from demonstrations toward exploratory resets. Once $x > 0.5$, sampling increasingly concentrates on feasible states while reducing exploratory resets.
            

\section{Experiments}
\label{sec:experiments}

    \begin{figure}[t]
        \small
        \centering
        \includegraphics[width=\textwidth]{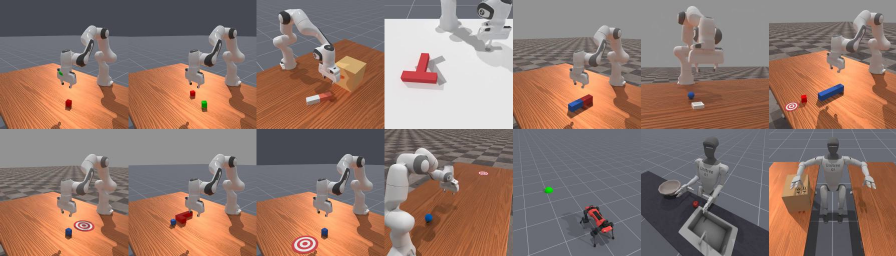}
        \caption{\textbf{Tasks.} Visualizations of the 14 tasks on which we evaluate FlashRFCL.}
        \label{fig:envs}
        \vspace{-10pt}
    \end{figure}

    We consider 14 challenging sparse-reward robotics tasks from ManiSkill3~\citep{taomaniskill3}: 11 tabletop tasks (PickCube, StackCube, PegInsertionSide, PushT, LiftPegUpright, PlaceSphere, PokeCube, PullCube, PullCubeTool, PushCube, and RollBall) and 3 quadruped/humanoid tasks (AnymalC-Reach, UnitreeG1PlaceAppleInBowl, UnitreeG1TransportBox), as shown in Fig. \ref{fig:envs}. As described in the past work ~\citep{tao2024rfcl, escoriza2025multistagemanipulationdemonstrationaugmentedreward}, these tasks are chosen as they are more difficult than most other benchmarks due to higher initial-state randomization, variation of the geometric properties of the object, and stricter precision requirements for success. All experiments are conducted on NVIDIA GeForce RTX 2080 Ti GPUs.

    \subsection{Main Results}

        \begin{figure}[t]
            \small
            \centering
            \includegraphics[width=\textwidth]{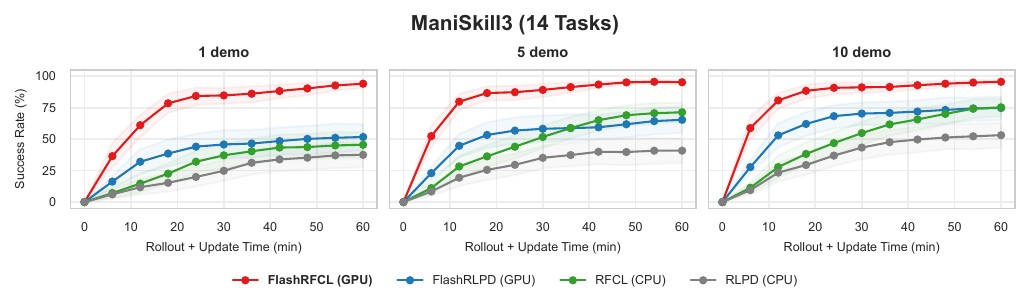}
            \vspace{-20pt}
            \caption{\textbf{Main Results.} Average success rate across all tasks given 1, 5, and 10 demonstrations. Shaded areas represent 95\% CIs over 5 seeds. The results show FlashRFCL is significantly faster than prior state-of-the-art performance, solving all tasks within one hour. FlashRFCL retains both asymptotic performance and wall-time efficiency even when only a single demonstration is provided.}
            \label{fig:main-results}
            \vspace{-10pt}
        \end{figure}
    
        We evaluate FlashRFCL against 3 state-of-the-art model-free RL baselines that use demonstrations and/or simulation states: RFCL, RLPD~\citep{ball2023efficientonlinereinforcementlearning}, and FlashRLPD. Flash RLPD is our implementation of RLPD leveraging GPU-based simulation, with the SAC backbone replaced by FlashSAC. Both RFCL and RLPD were originally setup to train with CPU simulations, we adapt them to use the GPU simulation but with far less parallel environments to emulate the low-parallelism of CPU simulation. Results in Fig. \ref{fig:main-results} shows that FlashRFCL is the only method that reaches near $100\%$ success rate within one hour when provided with a single demonstration. To the best of our knowledge, FlashRFCL is the first model-free RL method to solve all evaluated ManiSkill3 tasks from sparse rewards and a single demonstration. In contrast to FlashRFCL, all baselines exhibit higher sensitivity to the number of demonstrations, with the CPU simulation based algorithms being the most wall-time inefficient and needing far more demonstrations.
    
    \subsection{Training Without Demonstration Actions}
    \label{sec:train_without_demo_actions}
    \begin{figure}[h]
            \centering
            \vspace{-10pt}
            \small
            \includegraphics[width=0.75\textwidth]{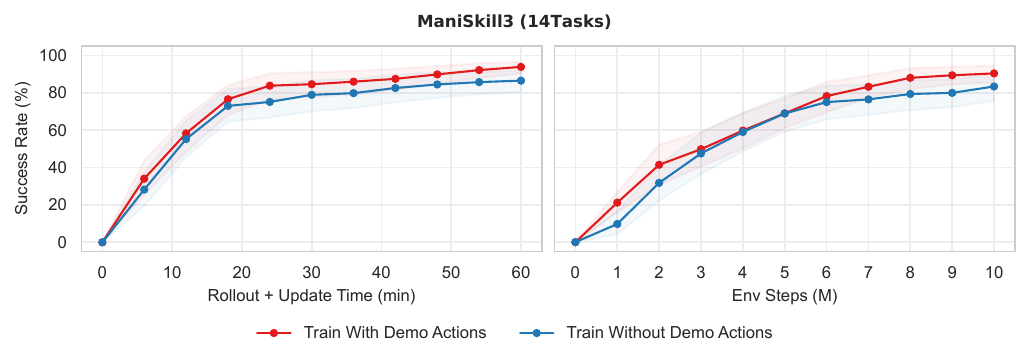}
            \vspace{-5pt}
            \caption{\textbf{Training Without Demonstration Actions.} Average success rate across all tasks when training with and without demonstration actions. Shaded areas represent 95\% CIs over 5 seeds. FlashRFCL is robust to demonstration action availability, with negligible difference in wall-time and sample efficiency.}
            \label{fig:train-without-demo-actions}
            \vspace{-10pt}
        \end{figure}

    Action labels are typically tightly coupled to how the actions were collected via teleoperation or scripting, whereas environment states are more reusable. Moreover, action labels may be noisy or suboptimal even when the demonstrated trajectory reaches success states. We test FlashRFCL on all tasks with and without demonstration actions, using a single demonstration per task. Without demonstration actions, we have no offline replay buffer and sample data entirely from the online replay buffer. Results in Fig. \ref{fig:train-without-demo-actions} demonstrate that FlashRFCL suffers only a small performance drop when we remove action labels, making FlashRFCL flexible in that it requires only simulation states and places no restrictions on how the actions used to obtain those simulation states are collected. For example, one could train FlashRFCL to use a joint-space controller while using data collected by a teleoperator who uses an end-effector pose controller.

\section{Analysis}
\label{sec:analysis}

    In this section, we analyze how different components of FlashRFCL impact training efficiency. All experiments are conducted on four tasks, namely PickCube, StackCube, PegInsertionSide, and PushT. These tasks are chosen because, together, they span four distinct and fundamental manipulation skills: pick-and-hold, precise stacking, non-prehensile pushing, and tight-clearance insertion.

    \subsection{Effect of Environment Parallelism on Reverse and Forward Curriculum}
        
        \begin{figure}[h]
            \centering
            \vspace{-10pt}
            \includegraphics[width=0.9\textwidth]{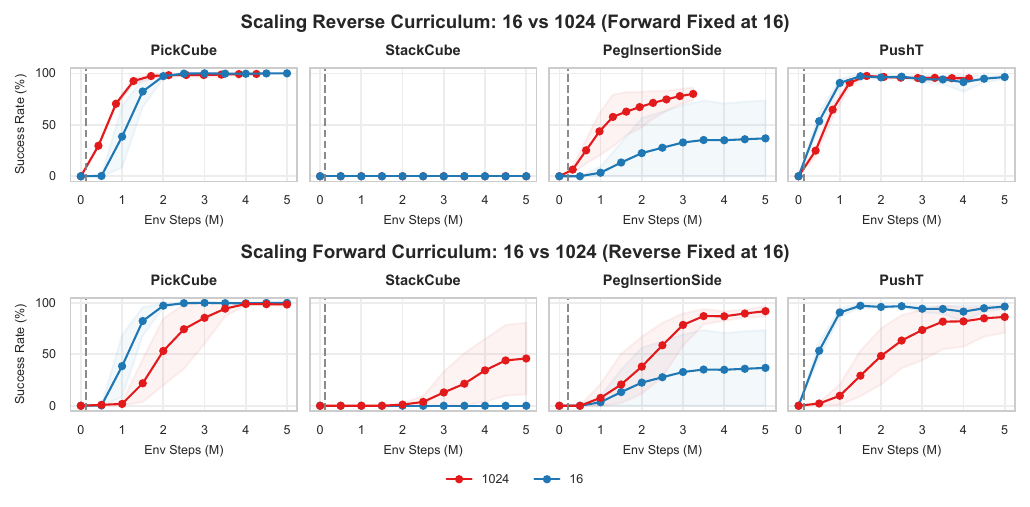}
            \vspace{-10pt}
            \caption{\textbf{Reverse to Forward.} Average success rate for each task given a single demonstration during the forward curriculum. Shaded areas represent 95\% confidence intervals over 5 seeds.}
            \label{fig:reverse-to-forward}
            \vspace{-10pt}
        \end{figure}

        We investigate the impact of environment parallelism with respect to the two curricula used in FlashRFCL. To isolate the effects of environment parallelism we disable DSS and construct $\Lambda_\text{train}$ by uniformly sampling initial states from $\mathcal{S}_\text{init}$ for the forward curriculum. Fig. \ref{fig:reverse-to-forward} shows the training curves of FlashRFCL during the forward curriculum when varying the number of parallel environments used during the reverse curriculum (top row) and during the forward curriculum (bottom row). To generate Fig. \ref{fig:reverse-to-forward}, we save a checkpoint once the reverse curriculum completes, then run forward curriculum training from the same checkpoint with different numbers of parallel environments.

        The top row shows that, with identical forward curriculum settings, trials that used 1024 parallel environments during the reverse curriculum were more sample-efficient on some tasks. This suggests that scaling up the reverse curriculum to more parallel environments can improve exploration and help learn representations that generalize better during the forward curriculum, particularly when exploration remains a bottleneck after reverse-solving a single demonstration.

        The bottom row shows that, starting from the same policy weights after the reverse curriculum, trials that used 1024 parallel environments during the forward curriculum achieved similar or higher success rates across all tasks after 5 million samples. With a fixed UTD ratio, increasing the number of parallel environments generally reduces sample efficiency but improves wall-time efficiency. Thus, when generalization from a single demonstration is relatively easy (PickCube, PushT), fewer environments yield higher sample efficiency. However, on challenging tasks where exploration remains a bottleneck after reverse-solving a single demonstration (StackCube, PegInsertionSide), increasing parallelism can mitigate exploration difficulties, facilitate forward curriculum progression, and even improve sample efficiency over using 16 environments. These results indicate that scaling up the forward curriculum can improve the generalization abilities of FlashRFCL, even when the initial policy has explored only states around a single demonstration during the reverse curriculum.

    \subsection{Forward Curriculum Ablations}

        \begin{figure}[t]
            \centering
            \vspace{-5pt}
            \includegraphics[width=0.9\textwidth]{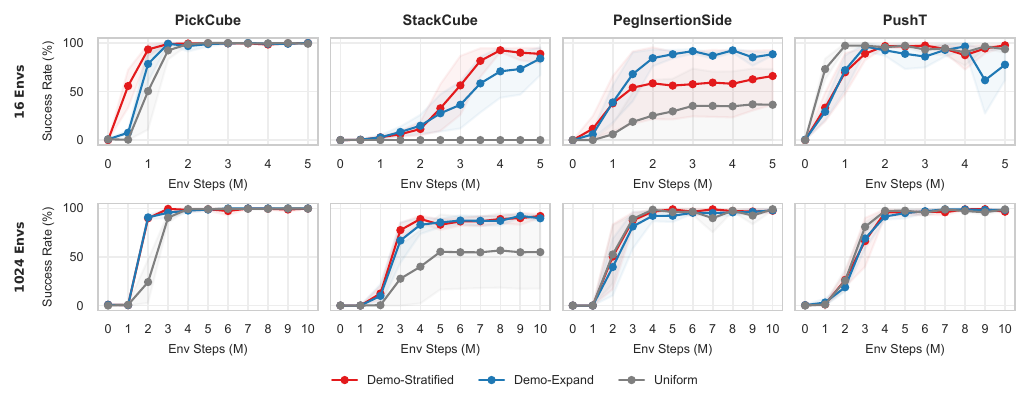}
            \vspace{-10pt}
            \caption{\textbf{Forward Curriculum Ablations.} Average success rate for each task given a single demonstration. Shaded areas represent 95\% CIs over 5 seeds. Demo-Stratified and Uniform denote Demo-Stratified and uniform sampling of $\Lambda_\text{train}$, respectively, with $\Lambda_\text{feasible}$ initialized as $\Lambda_\text{train}$. Demo-Expand denotes Demo-Stratified Sampling with $\Lambda_\text{feasible}$ initialized as $\Lambda_\text{demo} \cup \Lambda_\text{near-demo}^{(1)}$.}
            \label{fig:forward-ablations}
            \vspace{-10pt}
        \end{figure}

        We experiment with the forward curriculum by training FlashRFCL with 16 and 1024 environments. For each setting, we compare three variants: (1) forward curriculum with DSS while initializing $\Lambda_\text{feasible} = \Lambda_\text{train}$, (2) Demo-Expand, which uses DSS while initializing $\Lambda_\text{feasible} = \Lambda_\text{demo} \cup \Lambda_\text{near-demo}^{(1)}$, and (3) forward curriculum with uniform sampling of $\Lambda_\text{train}$ while initializing $\Lambda_\text{feasible} = \Lambda_\text{train}$. Fig. \ref{fig:forward-ablations} shows that training with 1024 environments is less sensitive to forward curriculum design choices than training with 16 environments. In contrast, with only 16 environments, uniform sampling of $\Lambda_\text{train}$ often leads to low success rates and fails entirely on StackCube. We find that DSS substantially improves learning with a small number of environments (e.g., 16) given only a single demonstration. Its benefits are task-dependent, however, and are most pronounced when exploration remains a bottleneck. For example, DSS can slightly reduce sample efficiency on relatively easier tasks such as PushT due to its bias toward demonstration-near initial states, but we observe no corresponding reduction in asymptotic success rates or wall-time efficiency. Overall, while uniform sampling works on most tasks, DSS improves the robustness of FlashRFCL across different levels of environment parallelism, particularly on more exploration-challenging tasks.
        
	\subsection{Base RL Algorithm Ablations}

        \begin{wrapfigure}{r}{0.5\linewidth}
            \centering
            \vspace{-2.0em}
            \includegraphics[width=0.95\linewidth]{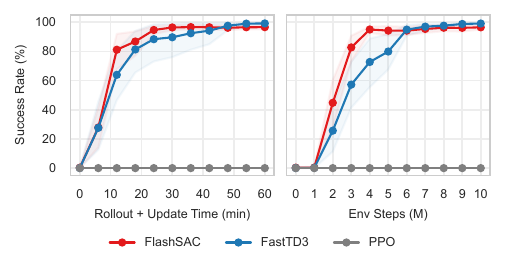}
            \vspace{-10pt}
            \caption{\textbf{Base RL Algorithm Ablations.} Average success rate of each base RL algorithm across all tasks given a single demonstration. Shaded areas represent 95\% CIs over 5 seeds. FlashSAC exhibits slightly better wall-time and sample efficiency than FastTD3. We find that PPO is not suitable for our setup.}
            \label{fig:algo-ablations}
            \vspace{-1.5em}
        \end{wrapfigure}
        
        We examine how the choice of base RL algorithm in FlashRFCL affects wall-time and sample efficiency. In addition to FlashSAC, we consider FastTD3 ~\citep{seo2025fasttd3} and PPO ~\citep{schulman2017proximalpolicyoptimizationalgorithms}, both of which are strong model-free RL algorithms well-suited for high-throughput GPU-based simulation. Because PPO is an on-policy algorithm, we use demonstrations only to reset the environment to states along the demonstration trajectories. As illustrated in Fig. \ref{fig:algo-ablations}, FlashSAC exhibits slightly better wall-time and sample efficiency over FastTD3. PPO on the other hand completely fails, indicating that the RFCL approach for sparse reward tasks with demonstrations is best used with off-policy algorithms.

    \subsection{Scaling Ablations}
    
        \begin{wrapfigure}{r}{0.45\linewidth}
            \centering
            \vspace{-2.0em}
            \includegraphics[width=1\linewidth]{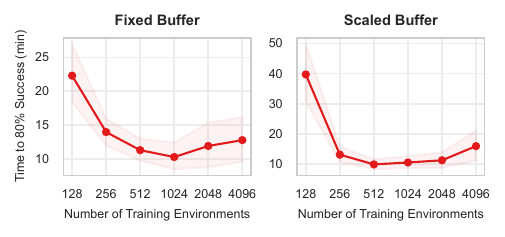}
            \vspace{-2em}
            \caption{\textbf{Env Scaling Ablations.} Wall-clock time to reach 80\% success rate vs number of environments. Results are aggregated across all tasks.}
            \label{fig:env-scaling-ablations}
            \vspace{-10pt}
        \end{wrapfigure}

        We analyze how scaling the number of environments affects the wall-time efficiency of FlashRFCL. To evaluate efficiency, we measure the wall-time required to reach a fixed performance threshold of 80\% success rate. We study scaling behavior under two configurations. In the fixed-buffer setting, we use a replay buffer size of 1M transitions for all numbers of environments. In the scaled-buffer setting, we set the buffer size to 1M for 1024 environments and scale it proportionally for other environment counts to maintain a consistent buffer refresh rate. In both configurations, we observe improvements in wall-clock efficiency up to a threshold of around 1024 environments. Specifically, wall-clock efficiency increases sharply at smaller scales, followed by diminishing returns as the number of environments increases. Beyond this threshold, further increasing the number of environments leads to degraded wall-clock efficiency. This suggests that the benefits of increased data diversity from scaling the number of environments no longer compensates for the additional rollout time due to non-linear scaling of GPU simulation throughput.

        \begin{figure}[t]
            \centering
            \vspace{-1em}
            \includegraphics[width=1.0\textwidth]{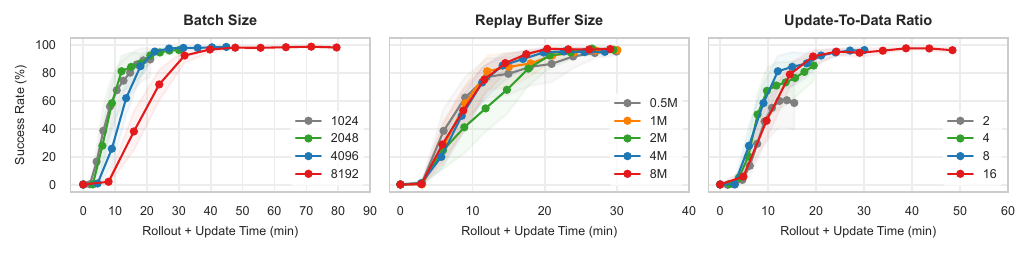}
            \vspace{-10pt}
            \caption{\textbf{Scaling Ablations.} We analyze the effects of scaling batch size, buffer size, and utd ratio. Shaded areas represent 95\% CIs over 5 seeds.}
            \label{fig:scaling-ablations}
            \vspace{-10pt}
        \end{figure}

        Similarly to prior work, we find that increasing the batch and replay buffer size increases training efficiency by providing a more stable learning signal. However, \Cref{fig:scaling-ablations} indicates that increasing the batch size beyond 2048 noticeably degrades wall-clock efficiency. In contrast to FlashSAC, which found that reducing the UTD ratio to 2/1024 incurs no degradation in wall-clock efficiency or performance, \Cref{fig:scaling-ablations} shows that overly reducing the UTD ratio in FlashRFCL can negatively impact wall-clock efficiency.
	

\section{Conclusion}
\label{sec:conclusion}

    We introduced FlashRFCL, a fast and demonstration-efficient learning-from-demonstration algorithm for high-throughput GPU simulation, capable of solving complex manipulation tasks from a single demonstration. We further introduced Demo-Stratified Sampling, which enhances the demonstration efficiency of FlashRFCL across both low- and high-parallelism settings. Through experiments, we show that FlashRFCL substantially improves on both wall-time efficiency and demonstration efficiency, and provide analysis on how parallelism impacts components of the algorithm.

	One limitation of FlashRFCL is reliance on resetting the environment to arbitrary states. Such resets are typically only feasible in simulation, which constrains FlashRFCL to simulated training environments. Consequently, real-world deployment of learned policy would require a sim-to-real transfer stage, but we consider this an orthogonal research direction to FlashRFCL and note that several past papers have demonstrated successful sim-to-real transfer using ManiSkill3 \citep{xie2026handelbotrealworldpianoplaying, fang2026dexdrummer, almuzairee2026squint}.


\section{Acknowledgments}
\label{sec:acknowledgments}

    Stone Tao is supported in part by the NSF Graduate Research Fellowship Program grant under grant No. DGE2038238.


\clearpage
\newpage


\bibliography{references}  

@misc{kim2026flashsacfaststableoffpolicy,
      title={FlashSAC: Fast and Stable Off-Policy Reinforcement Learning for High-Dimensional Robot Control}, 
      author={Donghu Kim and Youngdo Lee and Minho Park and Kinam Kim and I Made Aswin Nahendra and Takuma Seno and Sehee Min and Daniel Palenicek and Florian Vogt and Danica Kragic and Jan Peters and Jaegul Choo and Hojoon Lee},
      year={2026},
      eprint={2604.04539},
      archivePrefix={arXiv},
      primaryClass={cs.LG},
      url={https://arxiv.org/abs/2604.04539}, 
}

@misc{ball2023efficientonlinereinforcementlearning,
      title={Efficient Online Reinforcement Learning with Offline Data}, 
      author={Philip J. Ball and Laura Smith and Ilya Kostrikov and Sergey Levine},
      year={2023},
      eprint={2302.02948},
      archivePrefix={arXiv},
      primaryClass={cs.LG},
      url={https://arxiv.org/abs/2302.02948}, 
}

@inproceedings{10.1145/1553374.1553380,
author = {Bengio, Yoshua and Louradour, J\'{e}r\^{o}me and Collobert, Ronan and Weston, Jason},
title = {Curriculum learning},
year = {2009},
isbn = {9781605585161},
publisher = {Association for Computing Machinery},
address = {New York, NY, USA},
url = {https://doi.org/10.1145/1553374.1553380},
doi = {10.1145/1553374.1553380},
booktitle = {Proceedings of the 26th Annual International Conference on Machine Learning},
pages = {41–48},
numpages = {8},
location = {Montreal, Quebec, Canada},
series = {ICML '09}
}

@misc{jiang2021prioritizedlevelreplay,
      title={Prioritized Level Replay}, 
      author={Minqi Jiang and Edward Grefenstette and Tim Rocktäschel},
      year={2021},
      eprint={2010.03934},
      archivePrefix={arXiv},
      primaryClass={cs.LG},
      url={https://arxiv.org/abs/2010.03934}, 
}

@misc{parkerholder2023evolvingcurricularegretbasedenvironment,
      title={Evolving Curricula with Regret-Based Environment Design}, 
      author={Jack Parker-Holder and Minqi Jiang and Michael Dennis and Mikayel Samvelyan and Jakob Foerster and Edward Grefenstette and Tim Rocktäschel},
      year={2023},
      eprint={2203.01302},
      archivePrefix={arXiv},
      primaryClass={cs.LG},
      url={https://arxiv.org/abs/2203.01302}, 
}

@inproceedings{tao2024rfcl,
  title={Reverse Forward Curriculum Learning for Extreme Sample and Demonstration Efficiency in RL},
  author={Tao, Stone and Shukla, Arth and Chan, Tse-kai and Su, Hao},
  booktitle = {International Conference on Learning Representations (ICLR)},
  year={2024},
}

@article{seo2025fasttd3,
  author    = {Seo, Younggyo and Sferrazza, Carmelo and Geng, Haoran and Nauman, Michal and Yin, Zhao-Heng and Abbeel, Pieter},
  title     = {FastTD3: Simple, Fast, and Capable Reinforcement Learning for Humanoid Control},
  journal   = {arXiv:2505.22642},
  year      = {2025},
}

@misc{seo2025learningsimtorealhumanoidlocomotion,
      title={Learning Sim-to-Real Humanoid Locomotion in 15 Minutes}, 
      author={Younggyo Seo and Carmelo Sferrazza and Juyue Chen and Guanya Shi and Rocky Duan and Pieter Abbeel},
      year={2025},
      eprint={2512.01996},
      archivePrefix={arXiv},
      primaryClass={cs.RO},
      url={https://arxiv.org/abs/2512.01996}, 
}

@misc{li2023parallelqlearningscalingoffpolicy,
      title={Parallel $Q$-Learning: Scaling Off-policy Reinforcement Learning under Massively Parallel Simulation}, 
      author={Zechu Li and Tao Chen and Zhang-Wei Hong and Anurag Ajay and Pulkit Agrawal},
      year={2023},
      eprint={2307.12983},
      archivePrefix={arXiv},
      primaryClass={cs.LG},
      url={https://arxiv.org/abs/2307.12983}, 
}

@misc{haarnoja2018softactorcriticoffpolicymaximum,
      title={Soft Actor-Critic: Off-Policy Maximum Entropy Deep Reinforcement Learning with a Stochastic Actor}, 
      author={Tuomas Haarnoja and Aurick Zhou and Pieter Abbeel and Sergey Levine},
      year={2018},
      eprint={1801.01290},
      archivePrefix={arXiv},
      primaryClass={cs.LG},
      url={https://arxiv.org/abs/1801.01290}, 
}

@misc{schulman2017proximalpolicyoptimizationalgorithms,
      title={Proximal Policy Optimization Algorithms}, 
      author={John Schulman and Filip Wolski and Prafulla Dhariwal and Alec Radford and Oleg Klimov},
      year={2017},
      eprint={1707.06347},
      archivePrefix={arXiv},
      primaryClass={cs.LG},
      url={https://arxiv.org/abs/1707.06347}, 
}

@article{taomaniskill3,
  title={ManiSkill3: GPU Parallelized Robotics Simulation and Rendering for Generalizable Embodied AI},
  author={Stone Tao and Fanbo Xiang and Arth Shukla and Yuzhe Qin and Xander Hinrichsen and Xiaodi Yuan and Chen Bao and Xinsong Lin and Yulin Liu and Tse-kai Chan and Yuan Gao and Xuanlin Li and Tongzhou Mu and Nan Xiao and Arnav Gurha and Viswesh Nagaswamy Rajesh and Yong Woo Choi and Yen-Ru Chen and Zhiao Huang and Roberto Calandra and Rui Chen and Shan Luo and Hao Su},
  journal = {Robotics: Science and Systems},
  year={2025},
}

@misc{escoriza2025multistagemanipulationdemonstrationaugmentedreward,
      title={Multi-Stage Manipulation with Demonstration-Augmented Reward, Policy, and World Model Learning}, 
      author={Adrià López Escoriza and Nicklas Hansen and Stone Tao and Tongzhou Mu and Hao Su},
      year={2025},
      eprint={2503.01837},
      archivePrefix={arXiv},
      primaryClass={cs.LG},
      url={https://arxiv.org/abs/2503.01837}, 
}

@misc{openai2019dota2largescale,
      title={Dota 2 with Large Scale Deep Reinforcement Learning}, 
      author={OpenAI and : and Christopher Berner and Greg Brockman and Brooke Chan and Vicki Cheung and Przemysław D{\k{e}}biak and Christy Dennison and David Farhi and Quirin Fischer and Shariq Hashme and Chris Hesse and Rafal J{\'o}zefowicz and Scott Gray and Catherine Olsson and Jakub Pachocki and Michael Petrov and Henrique P. d. O. Pinto and Jonathan Raiman and Tim Salimans and Jeremy Schlatter and Jonas Schneider and Szymon Sidor and Ilya Sutskever and Jie Tang and Filip Wolski and Susan Zhang},
      year={2019},
      eprint={1912.06680},
      archivePrefix={arXiv},
      primaryClass={cs.LG},
      url={https://arxiv.org/abs/1912.06680}, 
}

@misc{openai2019solvingrubikscuberobot,
      title={Solving Rubik's Cube with a Robot Hand}, 
      author={OpenAI and Ilge Akkaya and Marcin Andrychowicz and Maciek Chociej and Mateusz Litwin and Bob McGrew and Arthur Petron and Alex Paino and Matthias Plappert and Glenn Powell and Raphael Ribas and Jonas Schneider and Nikolas Tezak and Jerry Tworek and Peter Welinder and Lilian Weng and Qiming Yuan and Wojciech Zaremba and Lei Zhang},
      year={2019},
      eprint={1910.07113},
      archivePrefix={arXiv},
      primaryClass={cs.LG},
      url={https://arxiv.org/abs/1910.07113}, 
}

@misc{rudin2022learningwalkminutesusing,
      title={Learning to Walk in Minutes Using Massively Parallel Deep Reinforcement Learning}, 
      author={Nikita Rudin and David Hoeller and Philipp Reist and Marco Hutter},
      year={2022},
      eprint={2109.11978},
      archivePrefix={arXiv},
      primaryClass={cs.RO},
      url={https://arxiv.org/abs/2109.11978}, 
}

@article{raffin2025isaacsim,
  title   = "Getting SAC to Work on a Massive Parallel Simulator: An RL Journey With Off-Policy Algorithms",
  author  = "Raffin, Antonin",
  journal = "araffin.github.io",
  year    = "2025",
  month   = "Feb",
  url     = "https://araffin.github.io/post/sac-massive-sim/"
}

@article{shukla2025fastsac,
  title   = "Speeding Up SAC with Massively Parallel Simulation",
  author  = "Shukla, Arth",
  journal = "https://arthshukla.substack.com",
  year    = "2025",
  month   = "Mar",
  url     = "https://arthshukla.substack.com/p/speeding-up-sac-with-massively-parallel"
}

@misc{nair2018overcomingexplorationreinforcementlearning,
      title={Overcoming Exploration in Reinforcement Learning with Demonstrations}, 
      author={Ashvin Nair and Bob McGrew and Marcin Andrychowicz and Wojciech Zaremba and Pieter Abbeel},
      year={2018},
      eprint={1709.10089},
      archivePrefix={arXiv},
      primaryClass={cs.LG},
      url={https://arxiv.org/abs/1709.10089}, 
}

@misc{florensa2018reversecurriculumgenerationreinforcement,
      title={Reverse Curriculum Generation for Reinforcement Learning}, 
      author={Carlos Florensa and David Held and Markus Wulfmeier and Michael Zhang and Pieter Abbeel},
      year={2018},
      eprint={1707.05300},
      archivePrefix={arXiv},
      primaryClass={cs.AI},
      url={https://arxiv.org/abs/1707.05300}, 
}

@inproceedings{yin2026omnireset,
  title={Emergent Dexterity via Diverse Resets and Large-Scale Reinforcement Learning},
  author={Patrick Yin and Tyler Westenbroek and Zhengyu Zhang and Joshua Tran and Ignacio Dagnino and Eeshani Shilamkar and Numfor Mbiziwo-Tiapo and Simran Bagaria and Xinlei Liu and Galen Mullins and Andrey Kolobov and Abhishek Gupta},
  booktitle={The Fourteenth International Conference on Learning Representations},
  year={2026},
  url={https://arxiv.org/abs/2603.15789},
}

@article{dextrahrgb,
  author       = {Ritvik Singh and
                  Arthur Allshire and
                  Ankur Handa and
                  Nathan D. Ratliff and
                  Karl Van Wyk},
  title        = {DextrAH-RGB: Visuomotor Policies to Grasp Anything with Dexterous
                  Hands},
  journal      = {CoRR},
  volume       = {abs/2412.01791},
  year         = {2024},
  url          = {https://doi.org/10.48550/arXiv.2412.01791},
  doi          = {10.48550/ARXIV.2412.01791},
  eprinttype    = {arXiv},
  eprint       = {2412.01791},
  bibsource    = {dblp computer science bibliography, https://dblp.org}
}

@article{DBLP:journals/corr/AmodeiOSCSM16,
  author       = {Dario Amodei and
                  Chris Olah and
                  Jacob Steinhardt and
                  Paul F. Christiano and
                  John Schulman and
                  Dan Man{\'{e}}},
  title        = {Concrete Problems in {AI} Safety},
  journal      = {CoRR},
  volume       = {abs/1606.06565},
  year         = {2016},
  url          = {http://arxiv.org/abs/1606.06565},
  eprinttype    = {arXiv},
  eprint       = {1606.06565},
  bibsource    = {dblp computer science bibliography, https://dblp.org}
}

@article{DBLP:journals/corr/abs-2511-14759,
  author       = {Physical Intelligence and
                  Ali Amin and
                  Raichelle Aniceto and
                  Ashwin Balakrishna and
                  Kevin Black and
                  Ken Conley and
                  Grace Connors and
                  James Darpinian and
                  Karan Dhabalia and
                  Jared DiCarlo and
                  Danny Driess and
                  Michael Equi and
                  Adnan Esmail and
                  Yunhao Fang and
                  Chelsea Finn and
                  Catherine Glossop and
                  Thomas Godden and
                  Ivan Goryachev and
                  Lachy Groom and
                  Hunter Hancock and
                  Karol Hausman and
                  Gashon Hussein and
                  Brian Ichter and
                  Szymon Jakubczak and
                  Rowan Jen and
                  Tim Jones and
                  Ben Katz and
                  Liyiming Ke and
                  Chandra Kuchi and
                  Marinda Lamb and
                  Devin LeBlanc and
                  Sergey Levine and
                  Adrian Li{-}Bell and
                  Yao Lu and
                  Vishnu Mano and
                  Mohith Mothukuri and
                  Suraj Nair and
                  Karl Pertsch and
                  Allen Z. Ren and
                  Charvi Sharma and
                  Lucy Xiaoyang Shi and
                  Laura Smith and
                  Jost Tobias Springenberg and
                  Kyle Stachowicz and
                  Will Stoeckle and
                  Alex Swerdlow and
                  James Tanner and
                  Marcel Torne and
                  Quan Vuong and
                  Anna Walling and
                  Haohuan Wang and
                  Blake Williams and
                  Sukwon Yoo and
                  Lili Yu and
                  Ury Zhilinsky and
                  Zhiyuan Zhou},
  title        = {{\(\pi\)}\({}^{\mbox{*}}\)\({}_{\mbox{0.6}}\): a {VLA} That Learns
                  From Experience},
  journal      = {CoRR},
  volume       = {abs/2511.14759},
  year         = {2025},
  url          = {https://doi.org/10.48550/arXiv.2511.14759},
  doi          = {10.48550/ARXIV.2511.14759},
  eprinttype   = {arXiv},
  eprint       = {2511.14759},
  bibsource    = {dblp computer science bibliography, https://dblp.org}
}

@misc{makoviychuk2021isaac,
      title={Isaac Gym: High Performance GPU-Based Physics Simulation For Robot Learning}, 
      author={Viktor Makoviychuk and Lukasz Wawrzyniak and Yunrong Guo and Michelle Lu and Kier Storey and Miles Macklin and David Hoeller and Nikita Rudin and Arthur Allshire and Ankur Handa and Gavriel State},
      year={2021},
      journal={arXiv preprint arXiv:2108.10470}
}

@inproceedings{todorov2012mujoco,
  title={MuJoCo: A physics engine for model-based control},
  author={Todorov, Emanuel and Erez, Tom and Tassa, Yuval},
  booktitle={2012 IEEE/RSJ International Conference on Intelligent Robots and Systems},
  pages={5026--5033},
  year={2012},
  organization={IEEE},
  doi={10.1109/IROS.2012.6386109}
}

@inproceedings{gail2016,
  author       = {Jonathan Ho and
                  Stefano Ermon},
  editor       = {Daniel D. Lee and
                  Masashi Sugiyama and
                  Ulrike von Luxburg and
                  Isabelle Guyon and
                  Roman Garnett},
  title        = {Generative Adversarial Imitation Learning},
  booktitle    = {Advances in Neural Information Processing Systems 29: Annual Conference
                  on Neural Information Processing Systems 2016, December 5-10, 2016,
                  Barcelona, Spain},
  pages        = {4565--4573},
  year         = {2016},
  url          = {https://proceedings.neurips.cc/paper/2016/hash/cc7e2b878868cbae992d1fb743995d8f-Abstract.html},
  bibsource    = {dblp computer science bibliography, https://dblp.org}
}

@article{lancaster2023modem,
  title={MoDem-V2: Visuo-Motor World Models for Real-World Robot Manipulation},
  author={Lancaster, Patrick and Hansen, Nicklas and Rajeswaran, Aravind and Kumar, Vikash},
  journal={arXiv preprint arXiv:2309.14236},
  year={2023}
}

@article{almuzairee2026squint,
  title = {Squint: Fast Visual Reinforcement Learning for Sim-to-Real Robotics},
  author = {Almuzairee, Abdulaziz and Christensen, Henrik I.},
  journal = {arXiv preprint arXiv:2602.21203},
  year = {2026},
}

@misc{fang2026dexdrummer,
      title={DexDrummer: In-Hand, Contact-Rich, and Long-Horizon Dexterous Robot Drumming}, 
      author={Hung-Chieh Fang and Amber Xie and Jennifer Grannen and Kenneth Llontop and Dorsa Sadigh},
      year={2026},
      eprint={2603.22263},
      archivePrefix={arXiv},
      primaryClass={cs.RO},
      url={https://arxiv.org/abs/2603.22263}, 
}

@misc{xie2026handelbotrealworldpianoplaying,
      title={HandelBot: Real-World Piano Playing via Fast Adaptation of Dexterous Robot Policies}, 
      author={Amber Xie and Haozhi Qi and Dorsa Sadigh},
      year={2026},
      eprint={2603.12243},
      archivePrefix={arXiv},
      primaryClass={cs.RO},
      url={https://arxiv.org/abs/2603.12243}, 
}

@article{goexplore,
  author       = {Adrien Ecoffet and
                  Joost Huizinga and
                  Joel Lehman and
                  Kenneth O. Stanley and
                  Jeff Clune},
  title        = {Go-Explore: a New Approach for Hard-Exploration Problems},
  journal      = {CoRR},
  volume       = {abs/1901.10995},
  year         = {2019},
  url          = {http://arxiv.org/abs/1901.10995},
  eprinttype   = {arXiv},
  eprint       = {1901.10995},
  bibsource    = {dblp computer science bibliography, https://dblp.org}
}

@inproceedings{tairanhe2025,
  author       = {Tairan He and
                  Wenli Xiao and
                  Toru Lin and
                  Zhengyi Luo and
                  Zhenjia Xu and
                  Zhenyu Jiang and
                  Jan Kautz and
                  Changliu Liu and
                  Guanya Shi and
                  Xiaolong Wang and
                  Linxi Jim Fan and
                  Yuke Zhu},
  title        = {{HOVER:} Versatile Neural Whole-Body Controller for Humanoid Robots},
  booktitle    = {{IEEE} International Conference on Robotics and Automation, {ICRA}
                  2025, Atlanta, GA, USA, May 19-23, 2025},
  pages        = {9989--9996},
  publisher    = {{IEEE}},
  year         = {2025},
  url          = {https://doi.org/10.1109/ICRA55743.2025.11128549},
  doi          = {10.1109/ICRA55743.2025.11128549},
  bibsource    = {dblp computer science bibliography, https://dblp.org}
}

@inproceedings{mu2024drs,

    title={DrS: Learning Reusable Dense Rewards for Multi-Stage Tasks},

    author={Mu, Tongzhou and Liu, Minghua and Su, Hao},

    booktitle={The Twelfth International Conference on Learning Representations},

    year={2024}

}

\newpage
\appendix
\section{Algorithmic Details}
\label{appendix:algo}
Algorithm \ref{alg:flashrfcl} provides pseudocode for DSS and FlashRFCL.

\begin{algorithm}[t!]
        \caption{FlashRFCL}
        \label{alg:flashrfcl}
        \begin{algorithmic}
            \footnotesize

            \LComment{\color{myblue} Demo-Stratified Sampling}
            \Function{DSS}{}
            \BeginBox[fill=white]
            \State Sample candidate states $\Lambda_\text{cand} \sim \mathcal S_\text{init}$
            \State Compute per-demonstration normalized pairwise distances (z-scores) between $\Lambda_\text{demo}$ and $\Lambda_\text{cand}$
            \State Partition $\Lambda_\text{cand}$ into strata $\{\Lambda_\text{near-demo}^{(k)}\}_{k=1}^{Q}$ using the quantiles
            \State Assign each state in $\Lambda_\text{cand}$ to a stratum using per-demonstration quantiles $\{q_k^{(i)}\}_{i=1, k=1}^{N, Q}$
            \State Construct $\Lambda_\text{near-demo}$ by uniformly sampling $N_k$ states from each stratum
            \State Sample $\Lambda_\text{uniform} \sim \mathcal S_\text{init}$
            \State $\Lambda_\text{train} \gets \Lambda_\text{demo} \cup \Lambda_\text{near-demo} \cup \Lambda_\text{uniform}$
            \EndBox
            \State \Return $\Lambda_\text{train}$
            \EndFunction

            \vspace{5pt}
            
            \LComment{\color{myblue} Training loop}
            \While{not converged}                
                \BeginBox[fill=white]
                \LComment{Rollout}
                \State Sample $a \sim \pi_\theta(\cdot|s)$ given the current state $s$, and take action $a$
                \State Observe next state $s'$ and reward $r$
                \State Store $(s, a, r, s')$ in replay buffer
                \EndBox
                \BeginBox[fill=white]
                    \LComment{Curriculum}
                \EndBox
                \If{episode from $s_{i, \text{init}}$ is done}
                    \If{$\text{rand}() < p_\text{reverse}$}
                        \State Sample new demo id $j$ and discrete offset value $k$
                        \State Sample $s_{j,t+k}$ from $\tau_j$ in $\mathcal{D}_{\text{demo}}$  
                    \Else                            
                        \If{$|\Lambda_\text{train} \setminus \Lambda_\text{feasible}| > 0$}
                            \State Add $s_{i, \text{init}}$ to $\Lambda_\text{feasible}$ if $\mathrm{return}(s_{i,\mathrm{init}}) > 0$
                            \State Sample $s_{j,\text{init}} \sim P_\text{demo-expand}(s \mid \Lambda_\text{train})$
                        \Else
                            \State $s_{j,\text{init}} \sim (1 - \rho) \cdot P_S(s_{j,\text{init}}|\Lambda_{\text{train}}, S) + \rho \cdot P_C(s_{j,\text{init}}|\Lambda_{\text{train}}, C, c)$
                        \EndIf
                    \EndIf
                \vspace{2pt}
                \State Reset environment to $s_{j,\text{init}}$
                \EndIf
                \vspace{5pt}
                \BeginBox[fill=white]
                \LComment{Update}
                \State Sample a replay batch from $\mathcal{D} \cup \mathcal{D}_{\mathrm{demo}}$ with offline sampling probability $\alpha \cdot p_{\text{reverse}}$
                \State Perform standard off-policy RL updates
                \EndBox
            \EndWhile
            \Return $\pi_\theta$
        
        \end{algorithmic}
    \end{algorithm}

\section{Reinforcement Learning }
\label{appendix:reinforcement_learning}
Table \ref{tab:hyperparameters} lists all hyperparemeters for FlashRFCL.
\begin{table}[h]
\small
\centering
    \begin{tabular}{ll}
    \hline
    \rowcolor{gray!30}
    \textbf{Hyperparameter} & \textbf{Value} \\
    \hline
    \rowcolor{gray!10}
    \textbf{FlashSAC} & \\
    \hline
    Discount factor $(\gamma)$ & 0.95 \\
    & (LiftPegUpright, PickCube, PlaceSphere, \\
    & PullCube, RollBall, StackCube) \\
    & 0.99 \\
    & (AnymalC-Reach, PegInsertionSide, PokeCube, \\
    & PullCubeTool, PushCube, PushT, \\
    & UnitreeG1PlaceAppleInBowl, UnitreeG1TransportBox) \\
    \hline
    \rowcolor{gray!10}
    \textbf{Common} & \\
    \hline
    Parallel environments & 1024 \\
    Replay buffer capacity & 1M \\
    Batch size & 2048 \\
    Update-to-data (UTD) ratio & 8/1024 \\
    TD steps (n-step) & 1 \\
    & (PushCube, StackCube, UnitreeG1TransportBox) \\
    & 3 \\
    & (AnymalC-Reach, LiftPegUpright, PegInsertionSide, \\
    & PickCube, PlaceSphere, PokeCube, \\
    & PullCube, PullCubeTool, PushT, \\
    & RollBall, UnitreeG1PlaceAppleInBowl) \\
    \hline
    \rowcolor{gray!10}
    \textbf{Actor} & \\
    \hline
    Number of blocks & 2 \\
    Hidden dimension & 128 \\
    Update delay & 2 \\
    \hline
    \rowcolor{gray!10}
    \textbf{Critic} & \\
    \hline
    Number of blocks & 2 \\
    Hidden dimension & 256 \\
    Target critic momemntum & 0.1 \\
    Number of critics & 2 \\
    Value prediction type & Categorical \\
    Categorical Support & [0, 1] \\
    Number of bins & 101 \\
    \hline
    \rowcolor{gray!10}
    \textbf{Temperature} & \\
    \hline
    Entropy target & 0.1 \\
    Initial value & 0.001 \\
    \hline
    \rowcolor{gray!10}
    \textbf{Noise Repeat} & \\
    \hline
    Zeta distribution exponent & 2 \\
    Maximum repeat limit & 16 \\
    \hline
    \rowcolor{gray!10}
    \textbf{Reverse Curriculum} & \\
    \hline
    Reverse Step Size ($\delta$) & 4 \\
    Per-trajectory Curriculum Buffer Size (m) & 10 \\
    Demonstration Horizon to Episode Horizon Ratio ($\phi$) & 3 \\
    \hline
    \rowcolor{gray!10}
    \textbf{Forward Curriculum} & \\
    \hline
    Forward Curriculum Score Threshold ($\omega$) & 0.75 \\
    Forward Curriculum Buffer Size (m) & 10 \\
    Forward Curriculum Score Temperature ($\beta$) & 0.1 \\
    Forward Curriculum Ranking Strategy & Rank \\
    Forward Curriculum Staleness Coefficient & 0.3 \\
    Forward Curriculum Number of Initial States ($n$) & 1024 \\
    \hline
    \rowcolor{gray!10}
    \textbf{Demo-Stratified Sampling} & \\
    \hline
    \rowcolor{blue!5} Number of candidate initial states ($m$) & $1024 \cdot 100$ \\
    \rowcolor{blue!5} Number of quantiles ($Q$) & 10 \\
    \rowcolor{blue!5} Exponential Decay Factor for Target Sampling Distribution & 0.75 \\
  
    \end{tabular}
    \caption{Hyperparameters for FlashRFCL. Hyperparameters introduced by this paper are highlighted in blue, which are for Demo-Stratified Sampling.}
\label{tab:hyperparameters}
\end{table}
\end{document}